\documentclass[letterpaper, 10 pt, conference]{ieeeconf}  
\IEEEoverridecommandlockouts                              
\usepackage{amsmath}
\usepackage{amsfonts}
\usepackage{booktabs} 
\usepackage{makecell}
\usepackage{graphicx}
\usepackage{color}
\usepackage{multirow}
\usepackage{pifont}
\usepackage{amsmath}
\usepackage{amssymb}
\usepackage{dsfont}
\usepackage{bm}
\usepackage{capt-of}
\definecolor{nbarrier}{RGB}{255, 120, 50}
\definecolor{nbicycle}{RGB}{255, 192, 203}
\definecolor{nbus}{RGB}{255, 255, 0}
\definecolor{ncar}{RGB}{0, 150, 245}
\definecolor{nconstruct}{RGB}{0, 255, 255}
\definecolor{nmotor}{RGB}{200, 180, 0}
\definecolor{npedestrian}{RGB}{255, 0, 0}
\definecolor{ntraffic}{RGB}{255, 240, 150}
\definecolor{ntrailer}{RGB}{135, 60, 0}
\definecolor{ntruck}{RGB}{160, 32, 240}
\definecolor{ndriveable}{RGB}{255, 0, 255}
\definecolor{nother}{RGB}{139, 137, 137}
\definecolor{nsidewalk}{RGB}{75, 0, 75}
\definecolor{nterrain}{RGB}{150, 240, 80}
\definecolor{nmanmade}{RGB}{213, 213, 213}
\definecolor{nvegetation}{RGB}{0, 175, 0}

\definecolor{nvcolor}{RGB}{119,185,0}
\definecolor{roadcolor}{RGB}{234,51,246}
\definecolor{sidewalkcolor}{RGB}{68,8,72}
\definecolor{parkingcolor}{RGB}{241,156,249}
\definecolor{othergroundcolor}{RGB}{160,32,76}
\definecolor{buildingcolor}{RGB}{246,202,69}
\definecolor{carcolor}{RGB}{111,149,238}
\definecolor{truckcolor}{RGB}{74,32,172}
\definecolor{bicyclecolor}{RGB}{136,227,242}
\definecolor{motorcyclecolor}{RGB}{37,59,146}
\definecolor{othervehiclecolor}{RGB}{96,81,242}
\definecolor{vegetationcolor}{RGB}{79, 173, 50}
\definecolor{trunkcolor}{RGB}{126, 65, 22}
\definecolor{terraincolor}{RGB}{171, 238, 105}
\definecolor{personcolor}{RGB}{234, 60, 49}
\definecolor{bicyclistcolor}{RGB}{234, 66, 195}
\definecolor{motorcyclistcolor}{RGB}{138, 42, 90}
\definecolor{fencecolor}{RGB}{238, 128, 69}
\definecolor{polecolor}{RGB}{252, 241, 161}
\definecolor{trafficsigncolor}{RGB}{233, 51, 35}
\definecolor{other-struct.color}{RGB}{255, 150, 0}
\definecolor{other-objectcolor}{RGB}{50, 255, 255}
\definecolor{lane-markingcolor}{RGB}{150, 255, 170}
\definecolor{color1}{RGB}{176, 36, 24}
\definecolor{color2}{RGB}{0, 176, 80}
\definecolor{color3}{RGB}{0, 0, 200}

\title{\LARGE \bf
VG3S: Visual Geometry Grounded Gaussian Splatting 
\\ for Semantic Occupancy Prediction
}

\author{Xiaoyang Yan$^*$, Muleilan Pei$^{*,\dagger}$, and Shaojie Shen
\thanks{\textbf{$^{*}$Equal Contribution}. \textbf{$^{\dagger}$Corresponding Author \& Project Lead}.}
\thanks{All authors are with the Department of Electronic and Computer Engineering, The Hong Kong University of Science and Technology, Hong Kong SAR, China. Email: {\texttt{\{xyanaq,mpei,eeshaojie\}@ust.hk}}}
}

\makeatletter
\let\@oldmaketitle\@maketitle
\renewcommand{\@maketitle}{%
    \@oldmaketitle
    \centering
    \setcounter{figure}{0}
    \vspace{0.4cm}
    \includegraphics[width=\textwidth]{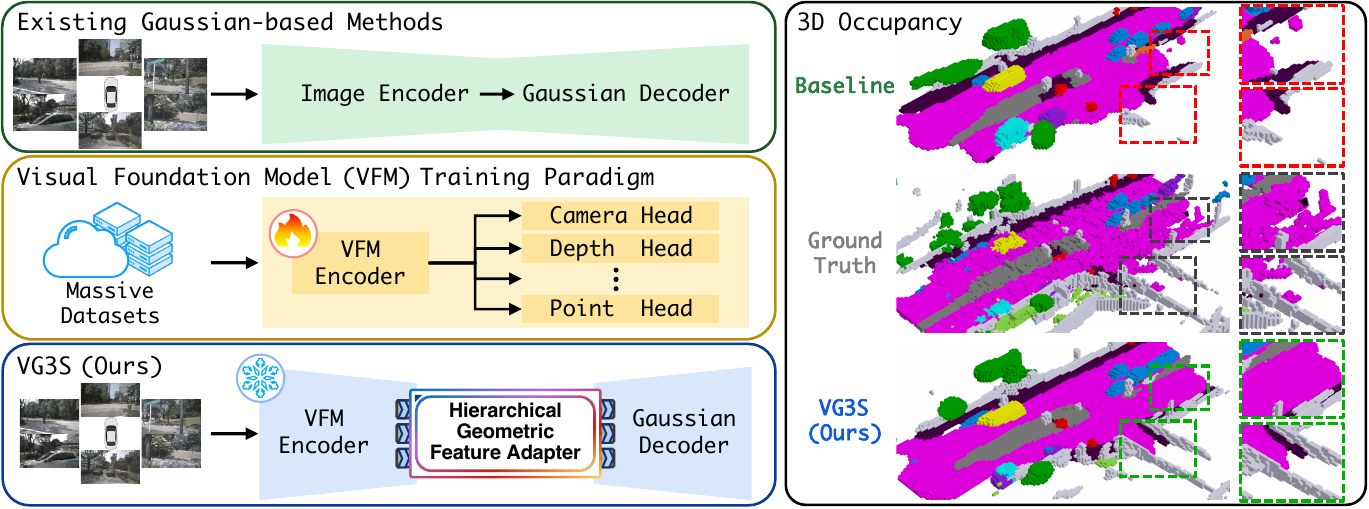}
    \captionof{figure}{\textbf{Comparison between existing Gaussian-based methods and our proposed VG3S.} Existing approaches often produce semantic occupancy with incomplete object coverage due to the lack of accurate 3D geometric priors. In contrast, our VG3S incorporates rich 3D geometric priors embedded in a frozen VFM pre-trained on massive datasets, enabling the decoder to generate more geometrically accurate and consistent semantic occupancy predictions.
    }
    \label{fig:fig1}
}
\makeatother

\begin{document}

\maketitle
\thispagestyle{empty}
\pagestyle{empty}

\begin{abstract}
3D semantic occupancy prediction has become a crucial perception task for comprehensive scene understanding in autonomous driving. While recent advances have explored 3D Gaussian splatting for occupancy modeling to substantially reduce computational overhead, the generation of high-quality 3D Gaussians relies heavily on accurate geometric cues, which are often insufficient in purely vision-centric paradigms. To bridge this gap, we advocate for injecting the strong geometric grounding capability from Vision Foundation Models (VFMs) into occupancy prediction. In this regard, we introduce \underline{V}isual \underline{G}eometry \underline{G}rounded \underline{G}aussian \underline{S}platting (VG3S), a novel framework that empowers Gaussian-based occupancy prediction with cross-view 3D geometric grounding. Specifically, to fully exploit the rich 3D geometric priors from a frozen VFM, we propose a plug-and-play hierarchical geometric feature adapter, which can effectively transform generic VFM tokens via feature aggregation, task-specific alignment, and multi-scale restructuring. Extensive experiments on the nuScenes occupancy benchmark demonstrate that VG3S achieves remarkable improvements of 12.6\% in IoU and 7.5\% in mIoU over the baseline. Furthermore, we show that VG3S generalizes seamlessly across diverse VFMs, consistently enhancing occupancy prediction accuracy and firmly underscoring the immense value of integrating priors derived from powerful, pre-trained geometry-grounded VFMs.

\end{abstract}

\section{Introduction}
3D semantic occupancy prediction has emerged as a cornerstone perception module for vision-centric autonomous driving systems \cite{pan2024renderocc, pei2025sept}, providing a dense volumetric representation that jointly encodes scene geometry and semantics. Unlike traditional 3D detection tasks \cite{detr3d, li2022stereo, li2025learning}, occupancy prediction yields a significantly more comprehensive understanding of the driving environment, which is indispensable for enabling safe navigation in complex urban scenarios.

Existing approaches typically model occupancy leveraging discrete voxels \cite{li2023voxformer, Riegler2017OctNet} or Bird's-Eye-View (BEV) representations \cite{li2022bevformer, pei2025goirl}, where multi-view image features are encoded and lifted into 3D space for dense occupancy forecasting. 
Recently, Gaussian-based scene formulations have become a promising alternative \cite{GaussianFormer, huang2025gaussianformer2}. By explicitly modeling a scene as a set of 3D Gaussian primitives \cite{kerbl3dgs} and rendering them via Gaussian-to-voxel splatting, these methods provide more interpretable scene representations while naturally exploiting spatial sparsity. This design avoids dense prediction over large empty regions, significantly improving memory efficiency while preserving fine-grained geometric details.

Despite these advancements, most models rely on feature extractors trained solely with limited occupancy supervision, which prevents them from learning strong 3D geometric priors and explicit cross-view constraints. As a result, these approaches often struggle to maintain structural consistency across views, leading to fragmented object geometries, such as incomplete drivable surface planes and man-made structures, as demonstrated in the top of Fig.~\ref{fig:fig1}. Consequently, the performance is heavily bottlenecked by the geometric understanding derived from sparse 3D supervision. Moreover, jointly optimizing image encoders with complex 2D-to-3D lifting modules often results in excessive training cycles and the potential loss of detailed 3D geometric information.

Meanwhile, recent geometry-grounded Visual Foundation Models (VFMs), such as VGGT \cite{wang2025vggt}, have exhibited strong cross-view geometric grounding capabilities through large-scale pre-training. As shown in the middle of Fig.~\ref{fig:fig1}, these models are trained on multi-view images with diverse 3D supervision signals under multi-task objectives. Such a training paradigm encourages the VFM to learn intrinsic correlations and viewpoint constraints among different geometric properties, thereby acquiring highly robust geometric consistency.

Unlike conventional occupancy frameworks that must infer 3D structure from limited supervision, these VFMs encode transferable geometric knowledge in their intermediate representations, including relative depth, structural boundaries, and multi-view correspondence. Such grounding priors offer valuable geometric information that can significantly benefit occupancy prediction, particularly for generalization to unseen environments. However, fully fine-tuning a VFM backbone is extremely resource-intensive, introduces a massive computational burden, and may risk catastrophic forgetting of the universal 3D geometric priors.

In light of the powerful geometric knowledge embedded in VFMs, we explore how to leverage these rich 3D cues from a frozen VFM to empower the Gaussian-based occupancy prediction framework. Nevertheless, directly applying raw features from the pre-trained VFMs for occupancy prediction remains challenging.
To bridge this gap, we introduce a plug-and-play Hierarchical Geometric Feature Adapter (HGFA) to transform generic VFM embeddings into occupancy-specific visual features, as illustrated at the bottom of Fig. \ref{fig:fig1}. Specifically, we utilize a grouped adaptive token fusion strategy to aggregate VFM tokens into a compact representation across hierarchical layers, suppressing redundant geometric activations. We then perform task-aligned token refinement to filter out task-irrelevant noise and calibrate the extracted geometric priors. Finally, we construct a latent spatial feature pyramid to enhance spatial modeling and enforce local geometric coherence in the feature space, producing structurally consistent multi-scale representations tailored for Gaussian-based decoding. Through this hierarchical adaptation pipeline, the HGFA effectively injects geometry-grounded VFM knowledge into the occupancy prediction pipeline, enabling the Gaussian-based decoder to generate more accurate, coherent, and structurally consistent semantic occupancy predictions.

In summary, our main contributions are as follows:
\begin{itemize}
\item We propose Visual Geometry Grounded Gaussian Splatting (VG3S), a flexible framework that empowers semantic occupancy prediction with superior cross-view geometric grounding derived from pre-trained VFMs.
\item We design a plug-and-play hierarchical geometric feature adapter that unleashes rich 3D geometric priors of frozen VFMs by injecting the generic VFM embeddings into visual tokens tailored for Gaussian-based decoding.
\item Extensive experiments on the nuScenes dataset validate that our  VG3S significantly improves occupancy prediction accuracy compared to the baseline and generalizes seamlessly across diverse geometry-grounded VFMs.
\end{itemize}

\section{Related Work}
\subsection{Scene-Centric Semantic Occupancy Prediction}
Semantic occupancy prediction has been receiving increasing attention as a scene-centric alternative to object-centric 3D detection, aiming to model the entire 3D environment with dense semantic labels rather than sparse instance-level bounding boxes. Most existing methods follow a 2D-to-3D lifting paradigm, projecting multi-view image features into a unified 3D representation for occupancy prediction. Voxel-based approaches, such as OccFormer~\cite{zhang2023occformer} and SurroundOcc~\cite{wei2023surroundocc}, construct volumetric grids from camera features to predict semantic occupancy. To reduce the high computational cost of dense voxelization, BEVFormer~\cite{li2022bevformer} adopts a BEV representation, while TPVFormer~\cite{tpvformer} further extends this idea by employing a tri-plane formulation to capture complementary spatial contexts. Despite their effectiveness, these grid-based formulations require exhaustive prediction across all cells in the discretized space, including large volumes of empty regions, resulting in substantial redundancy and limited scalability. In contrast, recent Gaussian-based 3D scene representations provide a more compact and structure-aware modeling paradigm by allocating computation only to occupied or relevant regions. This observation motivates our adoption of the Gaussian-based scene representation for semantic occupancy prediction.

\begin{figure*}[t]
    \centering
    \includegraphics[width=\textwidth]{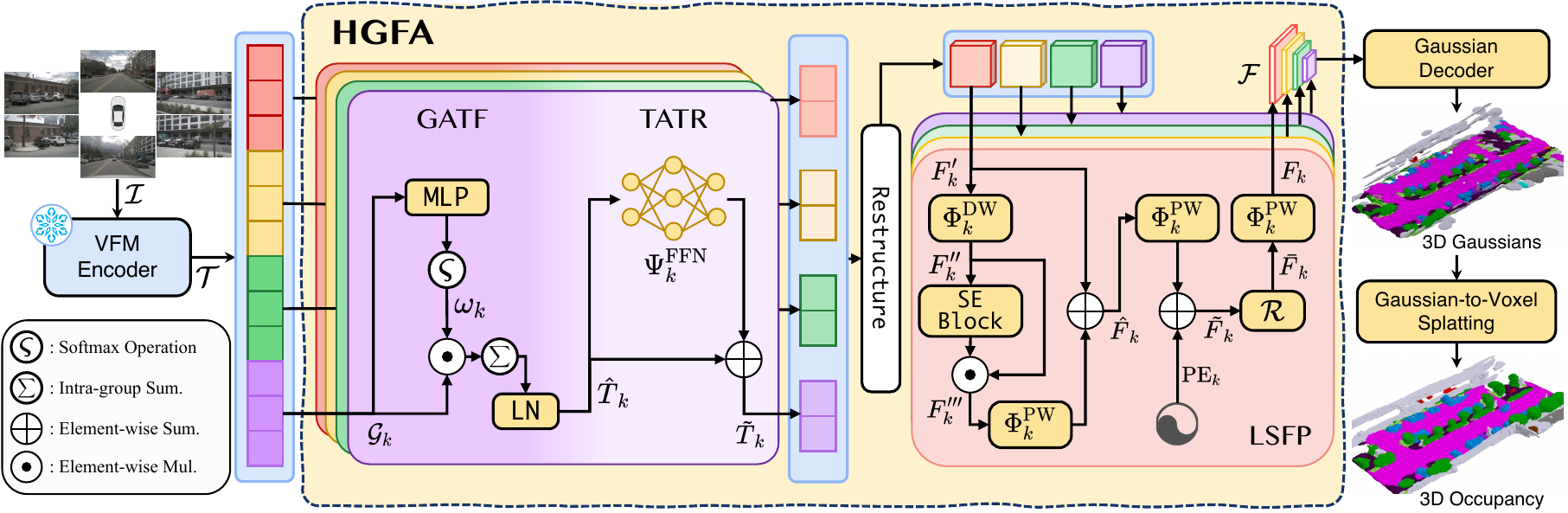}
    \caption{\textbf{Framework overview of VG3S}. Our approach leverages a powerful, pre-trained frozen VFM to provide rich 3D geometric priors, empowering the downstream Gaussian-based decoder with cross-view 3D geometric grounding and thereby significantly improving 3D semantic occupancy prediction.
    }
    \label{fig:fig2}
\end{figure*}

\subsection{Gaussian-Based 3D Scene Modeling}
3D Gaussian Splatting (3DGS)~\cite{kerbl3dgs} has recently advanced 3D scene reconstruction by introducing Gaussian primitives that enable efficient differentiable rendering and explicit spatial modeling. Compared to implicit representations such as Neural Radiance Fields (NeRF)~\cite{mildenhall2021nerf}, 3DGS offers real-time rendering capability, making it particularly well suited for large-scale outdoor scene understanding, including semantic occupancy prediction in autonomous driving. Building upon this paradigm, GaussianFormer~\cite{GaussianFormer} represents driving scenes using a set of sparse semantic 3D Gaussian primitives and performs occupancy prediction via Gaussian-to-voxel splatting. GaussianFormer-2~\cite{huang2025gaussianformer2} further enhances representation efficiency through distribution-based initialization and probabilistic Gaussian superposition, effectively reducing redundancy and excessive overlap among Gaussian primitives. Despite these promising efficiency gains, existing Gaussian-based methods still rely heavily on learnable image feature encoders trained with limited 3D supervision. As a result, the learned representations often lack robust geometric awareness, which degrades occupancy prediction performance. To address this limitation, our work incorporates rich geometry-grounded priors from visual foundation models to enhance Gaussian-based semantic occupancy prediction.

\subsection{Geometry-Grounded Visual Foundation Models}
Visual Foundation Models (VFMs) have exhibited strong performance and robust generalization across a broad spectrum of vision tasks. DINO~\cite{Dino} introduces self-distillation for self-supervised representation learning, enabling the discovery of semantic correspondences without human annotations. Subsequent models, including DINOv2~\cite{oquab2023dinov2} and DINOv3~\cite{simeoni2025dinov3}, demonstrate that large-scale pretraining on diverse web-scale datasets yields representations with strong semantic structure and cross-task generalization. Beyond semantic understanding, recent VFMs have further incorporated explicit geometric supervision. VGGT~\cite{wang2025vggt} leverages multi-task learning with depth estimation and camera pose prediction to extract geometry-grounded features directly from multi-view images. DVGT~\cite{zuo2025dvgt} and DGGT~\cite{dggt} also validate the effectiveness of such geometry-aware representations in autonomous driving, demonstrating robustness under dynamic and complex scenarios. Collectively, these studies underscore the potential of VFMs to offer consistent and geometry-grounded features across viewpoints. Concurrently, VG3T~\cite{kim2025vg3t} integrates VGGT into semantic occupancy prediction through full fine-tuning of the VFM backbone, which introduces significant computational overhead and risks undermining the generalization capability of pre-trained geometric priors. To this end, our work is intended to unleash the geometric grounding capability of VFMs without further tuning, thus preserving their generalization while effectively advancing Gaussian-based semantic occupancy prediction.

\section{Methodology}
\subsection{Problem Formulation}
The objective of 3D semantic occupancy prediction is to jointly infer the geometric occupancy and semantic category of a scene within a dense volumetric representation. Given a set of $S$-view images 
$\mathcal{I} = \{\mathcal{I}_{i} \in \mathbb{R}^{H \times W \times 3}\}_{i=1}^{S}$, 
where $H$ and $W$ denote the image height and width, the task is to predict a dense semantic occupancy grid $\mathcal{V} \in \mathcal{C}^{X \times Y \times Z}$, 
in which each voxel is assigned a semantic label from the category set $\mathcal{C}$. In view of the advantages of 3D Gaussian splatting in scene reconstruction~\cite{GaussianFormer, huang2025gaussianformer2}, we employ the 3D Gaussian representation for scene modeling. Specifically, the scene is represented by a set of $J$ Gaussian primitives 
$\mathcal{P} = \{\mathcal{P}_i\}_{i=1}^{J}$. 
Each primitive $\mathcal{P}_i$ is parameterized by a 3D position $\mathbf{m}_i \in \mathbb{R}^3$ and a covariance matrix $\mathbf{\Sigma}_i$ which is defined via a scaling vector $\mathbf{s}_i \in \mathbb{R}^3$ and a rotation quaternion $\mathbf{r}_i \in \mathbb{R}^4$. In addition, each Gaussian is associated with an opacity term $a_i \in [0,1]$ and a semantic logit vector $\mathbf{c}_i \in \mathbb{R}^{|\mathcal{C}|}$.
The final semantic occupancy prediction can be obtained via Gaussian-to-voxel splatting.

\subsection{Framework Overview}
The overall framework of VG3S is illustrated in Fig. \ref{fig:fig2}, which empowers semantic occupancy prediction with cross-view geometric grounding, thereby improving occupancy prediction performance. We first employ a pre-trained, frozen geometry-grounded VFM to extract visual tokens $\mathcal{T}$ from multi-view images $\mathcal{I}$. To effectively incorporate rich 3D geometric priors, we introduce a learnable Hierarchical Geometric Feature Adapter (HGFA) composed of three sequential modules: (i) Grouped Adaptive Token Fusion (GATF) for layer-wise feature aggregation, (ii) Task-Aligned Token Refinement (TATR) for goal-oriented feature calibration, and (iii) Latent Spatial Feature Pyramid (LSFP) for multi-scale feature restructuring. Through this hierarchical adaptation process, generic VFM embeddings are progressively transformed into geometry-enhanced tokens $\mathcal{F}$. These visual tokens are then decoded into Gaussian primitives, which are finally rendered into dense semantic occupancy voxels via a Gaussian-to-voxel splatting procedure.

\subsection{Geometry-Grounded VFM Feature Extraction}
Existing occupancy prediction approaches~\cite{GaussianFormer, huang2025gaussianformer2, wei2023surroundocc} typically rely on task-specific image encoders that are trained jointly with the downstream occupancy objective. However, due to the limited availability of large-scale 3D annotations, such dedicated feature extractors often lack robust cross-view geometric grounding capability and fail to generalize beyond the training distribution.
Recent advances in VFMs, pre-trained on diverse large-scale 3D datasets, have demonstrated strong 3D scene representation capacity. Motivated by this progress, we aim to inject such solid VFM-derived geometric priors into occupancy prediction to enhance cross-view geometric perception.

To preserve the universal geometric grounding capability of VFMs in open-world 3D environments while avoiding the substantial computational overhead of end-to-end VFM training, we employ the frozen VFM to encode surrounding images into visual features equipped with strong 3D geometric grounding consistency. Formally, let $\mathcal{E}$ denote the frozen VFM encoder. Given a surround-view image $\mathcal{I}_i$, the encoder $\mathcal{E}$ first divides it into $h \times w$ patches and encodes them into image tokens $T^{\mathrm{img}} \in \mathbb{R}^{L \times \mathrm{d}}$ using a DINO backbone~\cite{oquab2023dinov2, simeoni2025dinov3}, where $L = h \cdot w$ denotes the sequence length and $\mathrm{d}$ represents the embedding dimension of the DINO model:
\begin{equation}
T^{\mathrm{img}} = \mathcal{E}(\mathcal{I}_i).
\end{equation}
The image tokens are then augmented with camera tokens $T^{\mathrm{cam}} \in \mathbb{R}^{1 \times \mathrm{d}}$~\cite{darcet2023vision} and register tokens $T^{\mathrm{rgt}} \in \mathbb{R}^{4 \times \mathrm{d}}$~\cite{wang2024vggsfm}, yielding the augmented tokens $T \in \mathbb{R}^{(1 + 4 + L) \times \mathrm{d}}$:
\begin{equation}
T = T^{\mathrm{cam}} \boxplus T^{\mathrm{rgt}} \boxplus T^{\mathrm{img}},
\end{equation}
where $\boxplus$ denotes concatenation.

The augmented tokens $T$ are subsequently processed via a series of Alternating-Attention (AA) blocks~\cite{wang2025vggt}. Let $\mathcal{B}$ denote the ordered set of attention layers within each AA block. Herein, $\mathcal{B} = \{\mathrm{frame}, \mathrm{global}\}$ in VGGT~\cite{wang2025vggt} and DGGT~\cite{dggt}, while $\mathcal{B} = \{\mathrm{frame}, \mathrm{global}, \mathrm{temporal}\}$ in DVGT~\cite{zuo2025dvgt}. Across $N$ cascaded AA blocks, the initial augmented tokens are sequentially updated through these customized attention operations. Within the $j$-th block, the intermediate outputs of all $|\mathcal{B}|$ attention layers are concatenated to produce the updated tokens, denoted as $\mathrm{T}_j \in \mathbb{R}^{(1 + 4 + L) \times \mathcal{D}^{\mathrm{V}}}$, where $\mathcal{D}^{\mathrm{V}} = |\mathcal{B}|\cdot \mathrm{d}$. We retain the last $L$ image-relevant tokens as the final visual tokens $\mathcal{T}_j \in \mathbb{R}^{L \times \mathcal{D}^{\mathrm{V}}}$. Eventually, collecting the outputs from all blocks yields a set of geometry-grounded visual tokens, $\mathcal{T} = \{ \mathcal{T}_j \}_{j=1}^{N}$, which captures multi-level geometric priors of the 3D scene context.

\subsection{Hierarchical Geometric Feature Adapter}
Given the geometry-grounded visual token set $\mathcal{T}$ extracted from the frozen VFM, unleashing its rich geometric priors to the downstream Gaussian-based occupancy predictor remains a critical challenge. Direct utilization of these frozen tokens is suboptimal, as they lack the explicit spatial structure and task-specific alignment required for dense 3D forecasting. To bridge this inherent gap, we introduce a learnable Hierarchical Geometric Feature Adapter (HGFA), which effectively aggregates, aligns, and restructures the generic VFM features into geometry-enhanced representations tailored for semantic occupancy prediction.

\subsubsection{Grouped Adaptive Token Fusion}
We first partition the visual token set $\mathcal{T}$ into $K$ groups $\{\mathcal{G}_k \in \mathbb{R}^{M \times L \times \mathcal{D}^{\mathrm{V}}}\}_{k=1}^{K}$, where each group comprises visual tokens concatenated from $M = N/K$ consecutive layers that exhibit similar semantic granularity, i.e.,
\begin{equation}
\mathcal{G}_k = \mathcal{T}_{1+(k-1) \cdot M} \boxplus \mathcal{T}_{2+(k-1) \cdot M} \boxplus
\cdots \boxplus \mathcal{T}_{k \cdot M} .
\end{equation}
To effectively synthesize these intra-group tokens, we introduce the Grouped Adaptive Token Fusion (GATF) module, which leverages an adaptive feature fusion network to compute instance-specific combination weights. This design enables the dynamic suppression of redundant geometric responses across layers while selectively preserving the most informative scene features.

Formally, for the $k$-th group, we derive the layer-wise importance weights $\omega_{k}$ via:
\begin{equation}
\omega_{k} = \varsigma(\texttt{MLP}_{k}(\mathcal{G}_k)),
\end{equation}
where $\omega_{k} \in \mathbb{R}^{M \times L}$ represents the normalized importance scores across all $M$ layers within the group, $\varsigma(\cdot)$ denotes the Softmax function applied along the layer dimension, and $\texttt{MLP}(\cdot)$ is a multi-layer perceptron block.

The aggregated tokens $\hat{T}_{k} \in \mathbb{R}^{L \times \mathcal{D}^{\mathrm{V}}}$ for each group is then obtained through a weighted summation followed by normalization:
\begin{equation}
\hat{T}_{k} = \texttt{LN}_k\left( \sum_{m=1}^{M} \omega_{k,m} \odot \mathcal{G}_{k,m} \right),
\end{equation}
where $\texttt{LN}(\cdot)$ denotes layer normalization and $\odot$ represents element-wise multiplication.

\subsubsection{Task-Aligned Token Refinement}
Although the grouped visual tokens are compact, they still remain embedded within the generic latent manifold of the pre-trained VFM, which inevitably contains task-irrelevant activations. To bridge this gap and specialize the representations for the semantic occupancy prediction task, we introduce the Task-Aligned Token Refinement (TATR) module.

Designed as a streamlined residual block, TATR refines the generic VFM features into task-specific visual tokens with minimal computational overhead. Formally, the refined tokens $\tilde{T}_{k} \in \mathbb{R}^{L \times \mathcal{D}^{\mathrm{V}}}$ can be computed as:
\begin{equation}
\tilde{T}_{k} = \hat{T}_{k} + \Psi^{\mathrm{FFN}}_{k}(\hat{T}_{k}),
\end{equation}
where $\Psi^{\mathrm{FFN}}(\cdot)$ denotes a Feed-Forward Network (FFN) with GELU activation and dropout.

To balance representational capacity with efficiency, we equip the TATR module with a hierarchical capacity-scaling strategy. Recognizing that different groups contain scene information at varying levels of abstraction, we assign group-dependent hidden dimensions to their respective FFN. For the $k$-th group, the embedding dimension is expanded to $\rho_k \cdot \mathcal{D}^{\mathrm{V}}$, where $\rho_k$ is a group-specific expansion ratio. Larger expansion ratios are allocated to shallow groups to preserve fine-grained geometric details, whereas smaller ratios are used for deeper groups to distill compact, high-level semantics.

\subsubsection{Latent Spatial Feature Pyramid}
Having obtained the task-aligned visual tokens $\tilde{T}_{k}$, we first restore their spatial structure by reshaping them into grid-shaped features $F^{\prime}_{k} \in \mathbb{R}^{h \times w \times \mathcal{D}^{\mathrm{V}}}$, adhering to the original VFM patch layout. To reinforce local geometric coherence and establish robust spatial correspondences in the latent feature space, we introduce the Latent Spatial Feature Pyramid (LSFP) module.

Given $F_{k}^{\prime}$ as input, LSFP first applies depth-wise convolution to capture local spatial context:
\begin{equation}
F_{k}^{\prime \prime} = \Phi^{\mathrm{DW}}_{k}(F_{k}^{\prime}),
\end{equation}
where $\Phi^{\mathrm{DW}}(\cdot)$ denotes depth-wise convolution.

To further integrate the global contextual information, we incorporate a Squeeze-and-Excitation (SE) mechanism~\cite{hu2018senet}, which performs adaptive channel-wise reweighting:
\begin{equation}
F_{k}^{\prime \prime \prime} =
\sigma\!\left(\Psi_{k}^{\mathrm{FC}}(\xi(F_{k}^{\prime \prime}))\right)
\odot F_{k}^{\prime \prime},
\end{equation}
where $\xi(\cdot)$ denotes global average pooling, $\Psi^{\mathrm{FC}}(\cdot)$ represents fully connected layers, and $\sigma(\cdot)$ is the sigmoid activation.

A point-wise convolution is then utilized to facilitate cross-channel interaction and produce the refined spatial feature $\hat{F}_k \in \mathbb{R}^{h \times w \times \mathcal{D}^{\mathbb{V}}}$ with a residual connection:
\begin{equation}
\hat{F}_{k} = F_{k}^{\prime} + \Phi^{\mathrm{PW}}_{k}(F_{k}^{\prime \prime \prime}),
\end{equation}
where $\Phi^{\mathrm{PW}}(\cdot)$ denotes point-wise convolution.

Subsequently, to extract multi-scale features, we construct a feature pyramid across token groups, where each group is projected to a distinct expanded hidden dimension $\mathcal{D}_k^{\mathcal{H}}$ using pointwise convolution, followed by the injection of explicit spatial priors via a 2D sinusoidal positional embedding ~\cite{DPT}:
\begin{equation}
\tilde{F}_{k} = \Phi^{\mathrm{PW}}_{k}(\hat{F}_{k}) + \mathrm{PE}_k, 
\end{equation}
where $\mathrm{PE}_{k} \in \mathbb{R}^{ h \times w \times \mathcal{D}_k^{\mathcal{H}}}$ is the 2D positional embedding.

Next, we assign each group a distinct spatial scale factor $\tau_k$ and apply scale-adaptive convolutional resampling to $\tilde{F}_{k}$, yielding multi-resolution features $\bar{F}_{k} \in \mathbb{R}^{ (\tau_k \cdot h) \times (\tau_k \cdot w) \times \mathcal{D}_k^{\mathcal{H}}}$:
\begin{equation}
\bar{F}_{k} =  \mathcal{R}_k(\tilde{F}_{k}), 
\end{equation}
where $\mathcal{R}(\cdot)$ denotes a scale-specific resampling operator that adjusts the spatial resolution according to the factor $\tau_k$.

Each level of the feature pyramid is projected to the target channel dimension $\mathcal{D}$ aligned with the Gaussian decoder via point-wise convolution:
\begin{equation}
F_{k} = \Phi^{\mathrm{PW}}_{k} (\bar{F}_{k}), 
\end{equation}

Finally, the grid-shaped spatial features $\{F_{k}\}_{k=1}^K$ are flattened back into token sequences and concatenated together, yielding the ultimate geometry-grounded visual tokens $\mathcal{F} \in \mathbb{R}^{\left(L\cdot\sum_{k=1}^{K}(\tau_k)^2\right) \times \mathcal{D}}$ for 3D semantic occupancy prediction.

\subsection{Gaussian-to-Voxel Splatting}
Given the geometry-grounded visual tokens $\mathcal{F}$ produced by HGFA, we decode them into a set of semantic 3D Gaussian primitives $\mathcal{P}$. To fully exploit the cross-view geometric features embedded in $\mathcal{F}$, we adopt view-guided deformable attention \cite{li2024viewformer, yan2025stgs} for reference point sampling. Following the probabilistic Gaussian superposition framework~\cite{huang2025gaussianformer2}, the overall probability of occupancy can be obtained by aggregating the contributions of all Gaussian primitives. Semantic labels are then derived through a normalized expectation over Gaussian-conditioned class prediction. Finally, the geometry and semantics predictions are combined to generate the dense semantic occupancy prediction.

\subsection{Training Objective}
We optimize the network using a weighted combination of the standard cross-entropy loss $\mathcal{L}_{\mathrm{CE}}$ and the Lov\'{a}sz-Softmax loss $\mathcal{L}_{\mathrm{Lov}}$ \cite{berman2018lovaszloss}, as commonly adopted in semantic occupancy prediction \cite{GaussianFormer, tpvformer}. The overall training objective $\mathcal{L}_{\mathrm{total}}$ is defined as:
\begin{equation}
\mathcal{L}_{\mathrm{total}} = \lambda \mathcal{L}_{\mathrm{CE}} + \beta \mathcal{L}_{\mathrm{Lov}},
\end{equation}
where $\lambda$ and $\beta$ are coefficients for balancing the two losses.

\begin{table*}[t]
    \caption{3D semantic occupancy prediction results on the nuScenes benchmark.} 
    \small
    \setlength{\tabcolsep}{0.005\linewidth}  
    \vspace{-1.5mm}  
    \renewcommand\arraystretch{1.05}
    \centering
    \resizebox{\textwidth}{!}{
    \begin{tabular}{l| r | c c | c c c c c c c c c c c c c c c c}
        \toprule
        Method & \multicolumn{1}{c|}{Venue} 
        & \makecell{SC \\ IoU} & \makecell{SSC \\ mIoU}
        & \rotatebox{90}{\textcolor{nbarrier}{$\blacksquare$} barrier}
        & \rotatebox{90}{\textcolor{nbicycle}{$\blacksquare$} bicycle}
        & \rotatebox{90}{\textcolor{nbus}{$\blacksquare$} bus}
        & \rotatebox{90}{\textcolor{ncar}{$\blacksquare$} car}
        & \rotatebox{90}{\textcolor{nconstruct}{$\blacksquare$} const. veh.}
        & \rotatebox{90}{\textcolor{nmotor}{$\blacksquare$} motorcycle}
        & \rotatebox{90}{\textcolor{npedestrian}{$\blacksquare$} pedestrian}
        & \rotatebox{90}{\textcolor{ntraffic}{$\blacksquare$} traffic cone}
        & \rotatebox{90}{\textcolor{ntrailer}{$\blacksquare$} trailer}
        & \rotatebox{90}{\textcolor{ntruck}{$\blacksquare$} truck}
        & \rotatebox{90}{\textcolor{ndriveable}{$\blacksquare$} drive. suf.}
        & \rotatebox{90}{\textcolor{nother}{$\blacksquare$} other flat}
        & \rotatebox{90}{\textcolor{nsidewalk}{$\blacksquare$} sidewalk}
        & \rotatebox{90}{\textcolor{nterrain}{$\blacksquare$} terrain}
        & \rotatebox{90}{\textcolor{nmanmade}{$\blacksquare$} manmade}
        & \rotatebox{90}{\textcolor{nvegetation}{$\blacksquare$} vegetation} \\
        \midrule
        MonoScene~\cite{cao2022monoscene} & CVPR 2022 & 23.96 & 7.31 & 4.03 &	0.35& 8.00& 8.04&	2.90& 0.28& 1.16&	0.67&	4.01& 4.35&	27.72&	5.20& 15.13&	11.29&	9.03&	14.86 \\
        Atlas~\cite{murez2020atlas} & ECCV 2020 & 28.66 & 15.00 & 10.64&	5.68&	19.66& 24.94& 8.90&	8.84&	6.47& 3.28&	10.42&	16.21&	34.86&	15.46&	21.89&	20.95&	11.21&	20.54 \\
        BEVFormer~\cite{li2022bevformer} & ECCV 2022 & 30.50 & 16.75 & 14.22 &	6.58 & 23.46 & 28.28& 8.66 &10.77& 6.64& 4.05& 11.20&	17.78 & 37.28 & 18.00 & 22.88 & 22.17 & 13.80 &	\underline{22.21}\\
        TPVFormer~\cite{tpvformer} & CVPR 2023 & 11.51 & 11.66 & 16.14&	7.17& 22.63	& 17.13 & 8.83 & 11.39 & 10.46 & 8.23&	9.43 & 17.02 & 8.07 & 13.64 & 13.85 & 10.34 & 4.90 & 7.37\\
        TPVFormer$^\dagger$\cite{tpvformer}  & CVPR 2023 & {30.86} & 17.10 & 15.96&	 5.31& 23.86	& 27.32 & 9.79 & 8.74 & 7.09 & 5.20& 10.97 & 19.22 & {38.87} & {21.25} & {24.26} & {23.15} & 11.73 & 20.81\\
        OccFormer~\cite{zhang2023occformer} & ICCV 2023 & {31.39} & {19.03} & {18.65} & {10.41} & {23.92} & \underline{30.29} & {10.31} & {14.19} & \underline{13.59} & {10.13} & {12.49} & {20.77} & {38.78} & 19.79 & 24.19 & 22.21 & {13.48} & {21.35}\\
        SurroundOcc~\cite{wei2023surroundocc} & ICCV 2023 & \underline{31.49} & \underline{20.30}  & \underline{20.59} & {11.68} & \underline{28.06} & \textbf{30.86} & {10.70} & {15.14} & {\textbf{14.09}} & {\textbf{12.06}} & \underline{14.38} & \textbf{22.26} & 37.29 & \underline{23.70} & {24.49} & {22.77} & {\underline{14.89}} & {21.86}  \\
        GaussianFormer~\cite{GaussianFormer} & ECCV 2024 & 29.83 & {19.10} & {19.52} & {11.26} & {26.11} & {29.78} & {10.47} & {13.83} & {12.58} & {8.67} & {12.74} & {21.57} & {39.63} & {23.28} & {24.46} & {22.99} & 9.59 & 19.12 \\
        GaussianFormer-2*\cite{huang2025gaussianformer2} & CVPR 2025 & 30.56 & 20.02 & 20.15 & \textbf{12.99} & 27.61 & 30.23 & \underline{11.19} & \underline{15.31} & 12.64 & 9.63 & 13.31 & \textbf{22.26} & \underline{39.68} & 23.47 & \underline{25.62} & \underline{23.20} & 12.25 & 20.73 \\
        \midrule
        \textbf{VG3S (Ours)} & IROS 2026 &  \textbf{34.41} & \textbf{21.52} & \textbf{20.78} & \underline{12.40} & \textbf{28.09} & 29.53 & \textbf{11.93} & \textbf{15.59} & 12.38 & \underline{10.61} & \textbf{14.65} & \underline{21.74} & \textbf{42.42} & \textbf{26.39} & \textbf{28.06} & \textbf{26.58} & \textbf{17.46} & \textbf{25.76}  \\
        \bottomrule
    \end{tabular}}
    \begin{flushleft}{~~$\dagger$ denotes supervision using dense occupancy annotations \cite{wei2023surroundocc}. \\
    ~~* indicates results obtained with a 128-channel feature dimension for fair comparison \cite{huang2025gaussianformer2}.}\end{flushleft}
    \label{tab:tab1}
    \vspace{-0.5cm}
\end{table*}

\section{Experiments and Results}
\subsection{Experimental Setup}
\subsubsection{Dataset}
We conduct all experiments on the nuScenes dataset \cite{nuscenes}, which provides multi-view visual inputs from six synchronized cameras, collectively covering a $360^{\circ}$ horizontal field of view. The dataset comprises 1,000 urban driving sequences, each approximately 20 seconds long, with annotations provided at 2 Hz. In alignment with established benchmarks \cite{GaussianFormer, huang2025gaussianformer2}, we adopt a 700/150/150 sequence split for the training, validation, and testing sets, respectively. For supervision, we utilize dense semantic occupancy labels from SurroundOcc \cite{wei2023surroundocc} as ground truth. The target 3D space is defined as a $100\,\mathrm{m} \times 100\,\mathrm{m} \times 8\,\mathrm{m}$ volume centered at the ego vehicle, corresponding to a range of $[-50\,\mathrm{m}, 50\,\mathrm{m}]$ along both the $X$ and $Y$ axes and $[-5\,\mathrm{m}, 3\,\mathrm{m}]$ along the $Z$ axis. This volume is uniformly discretized into a $200 \times 200 \times 16$ voxel grid, resulting in a spatial resolution of $0.5\,\mathrm{m}$ per voxel.

\subsubsection{Metrics}
Following established evaluation protocols for semantic occupancy prediction \cite{cao2022monoscene}, we assess the performance of our proposed approach using two primary metrics based on the Intersection-over-Union (IoU). For the Semantic Scene Completion (SSC) task, which requires jointly predicting both scene geometry and semantic labels, we calculate the mean IoU (mIoU) of SSC across all occupied semantic categories $\mathcal{C}_{o}$:
\begin{equation}
\mathrm{mIoU} = \frac{1}{|\mathcal{C}_{o}|} \sum_{i \in \mathcal{C}_{o}} \frac{\mathrm{TP}_i}{\mathrm{TP}_i + \mathrm{FP}_i + \mathrm{FN}_i},
\end{equation}
where $\mathrm{TP}$, $\mathrm{FP}$, and $\mathrm{FN}$ denote the numbers of true positives, false positives, and false negatives, respectively.
In contrast, the Scene Completion (SC) task evaluates geometric reconstruction alone by ignoring semantic distinctions. All non-empty voxels are treated as a single occupied class $o$, and the corresponding IoU of SC is computed as:
\begin{equation}
\mathrm{IoU} = \frac{\mathrm{TP}_o}{\mathrm{TP}_o + \mathrm{FP}_o + \mathrm{FN}_o}.
\end{equation}

\subsubsection{Implementation Details}
We integrate the pre-trained geometry-grounded VFM, for example, DVGT \cite{zuo2025dvgt}, into the Gaussian-based occupancy framework \cite{huang2025gaussianformer2} for cross-view image feature extraction. The VFM remains entirely frozen throughout training, with its pre-trained weights kept fixed in all experiments.
For feature adaptation, the GATF module partitions tokens into $K=4$ groups, each containing $M=6$ layers. In TATR, the group-specific expansion ratios are set to $\{\rho_k\}_{k=1}^4 = \{4, 3, 2, 1.5\}$. In LSFP, the expanded hidden dimensions are set to $\{\mathcal{D}_k^{\mathcal{H}}\}_{k=1}^4 = \{768, 512, 384, 256\}$ with corresponding spatial scale factors $\{\tau_k\}_{k=1}^{4} = \{4, 2, 1, 0.5\}$.
Following established Gaussian-based designs \cite{GaussianFormer, huang2025gaussianformer2}, the decoder comprises four stacked transformer blocks utilizing $J=25,600$ Gaussian primitives, with the channel dimension $D$ fixed at 128. Training employs a cosine annealing learning rate schedule, preceded by a 500-iteration linear warm-up to a peak learning rate of $2 \times 10^{-4}$. To reduce the overfitting issue, we apply standard data augmentations, including image resizing and photometric distortions.

\begin{figure*}[t]
    \centering
    \includegraphics[width=\textwidth]{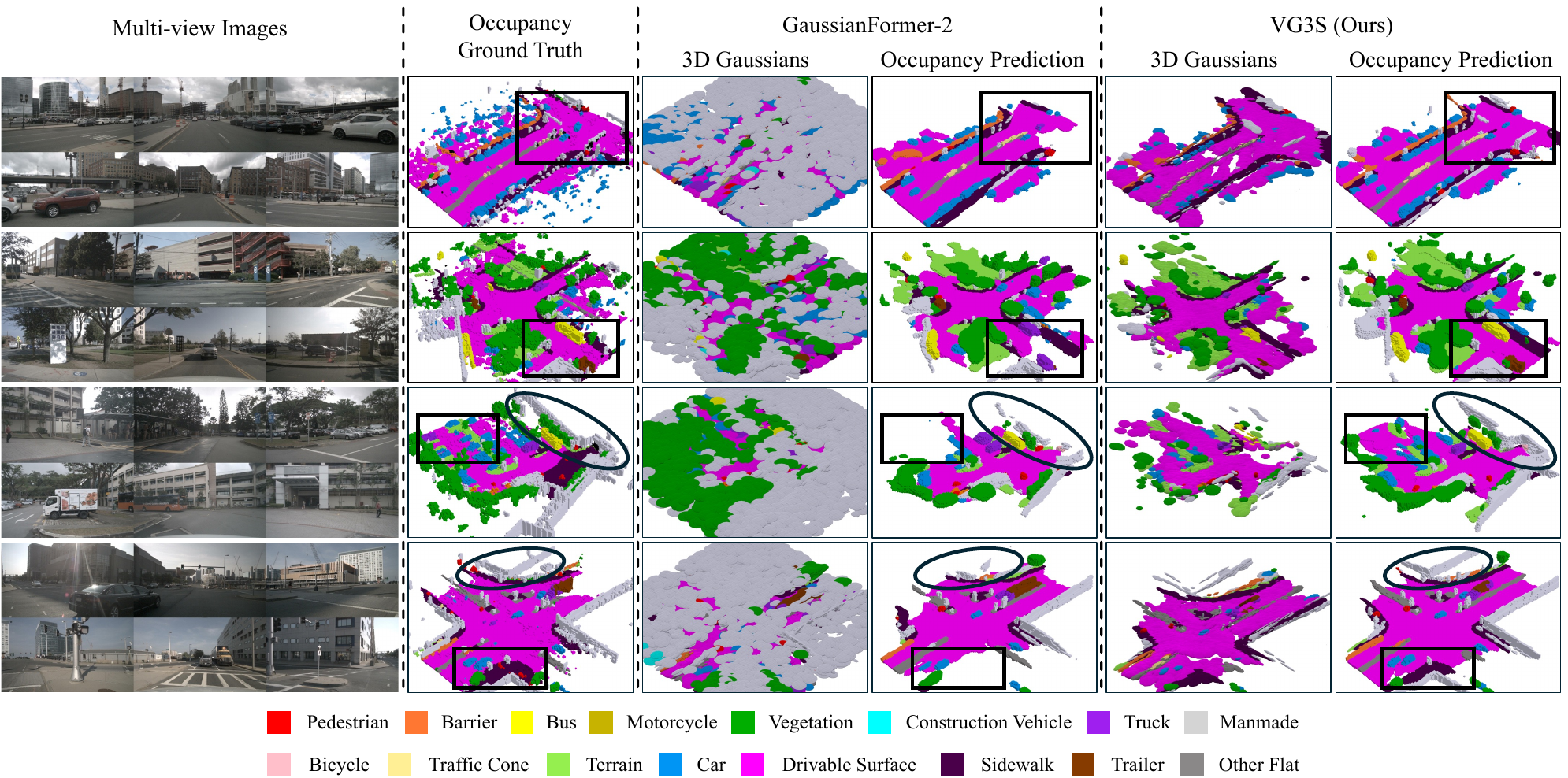}
    \caption{\textbf{Qualitative comparison between the baseline GaussianFormer-2 \cite{huang2025gaussianformer2} and our proposed VG3S}. Our approach produces more geometrically accurate and consistent object structures across four challenging scenes compared to the baseline, demonstrating that leveraging strong 3D geometric priors embedded within VFMs significantly improves 3D semantic occupancy predictions.
    }
    \label{fig:fig3}
    \vspace{-0.35cm}
\end{figure*}

\subsection{Quantitative Results}
\subsubsection{Comparison with State-of-the-Art}
The results on the nuScenes occupancy prediction benchmark are summarized in Table \ref{tab:tab1}, with the best and second-best scores indicated in \textbf{bold} and \underline{underlined}, respectively. Our proposed VG3S significantly outperforms the baseline, GaussianFormer-2 \cite{huang2025gaussianformer2}, achieving gains of 12.6\% in IoU and 7.5\% in mIoU. Furthermore, VG3S surpasses both prior voxel-based and existing Gaussian-based paradigms, demonstrating superior accuracy across most semantic classes. Notably, it yields compelling enhancements in structural categories, such as drivable surfaces, man-made objects, and vegetation. These findings not only validate the effectiveness of our proposed VG3S but also underscore that the rich 3D geometric priors derived from the frozen geometry-grounded VFM (DVGT \cite{zuo2025dvgt}) fundamentally enhance semantic occupancy prediction.

\subsubsection{Generalization Across Foundation Models}
We further investigate the generalizability and compatibility of our proposed framework by integrating it with multiple pre-trained VFMs. To ensure a comprehensive evaluation, we consider a diverse set of models with different design philosophies. Specifically, we include VGGT \cite{wang2025vggt}, which introduces multiple geometry-oriented training objectives built upon the DINOv2 backbone \cite{oquab2023dinov2}, and DGGT \cite{dggt}, which is fine-tuned on autonomous driving datasets. In addition, we incorporate the recent foundation model DINOv3 \cite{simeoni2025dinov3} and its geometry-aware variant DVGT \cite{zuo2025dvgt}, which is further adapted to driving scenarios.
As presented in Table \ref{tab:tab2}, integrating any of these VFMs into our pipeline consistently yields a substantial performance gain over the baseline (GaussianFormer-2), demonstrating the broad applicability of our design. Comparing VG3S equipped with DINOv2 (VG3S-DINOv2) against the geometric variants (VG3S-DGGT and VG3S-VGGT) reveals that although a stronger image backbone provides marginal benefits, the primary performance gains stem directly from the injected 3D geometric priors. Moreover, VG3S-DINOv3 achieves highly competitive semantic scene completion performance (21.36\% in mIoU), attributed to its robust, universal semantic representation capabilities. Ultimately, VG3S-DVGT, which synergizes cross-view 3D  geometric grounding with domain-specific autonomous driving knowledge, pushes the performance boundaries furthest, securing state-of-the-art scores of 34.41\% in IoU and 21.52\% in mIoU. These results confirm that our design is highly generalizable across diverse VFMs and exceptionally effective at exploiting their rich 3D geometric priors for 3D semantic occupancy predictions.

\begin{table}[t]
    \centering
    \caption{Evaluation of generalization across diverse VFMs.}
    \begin{tabular}{l|ccc}
        \toprule
        Method  & IoU &  mIoU  \\
        \midrule
        GaussianFormer-2 \cite{huang2025gaussianformer2} & 30.56 & 20.02 \\
        \midrule
        VG3S-DINOv2 \cite{oquab2023dinov2} & 32.34 &  20.42 \\ 
        VG3S-VGGT \cite{wang2025vggt} & 33.29 & 21.10 \\
        VG3S-DGGT \cite{dggt} & 33.37 & 20.81 \\
        \midrule
        VG3S-DINOv3 \cite{simeoni2025dinov3} & 33.20 & 21.36 \\
        VG3S-DVGT \cite{zuo2025dvgt} & \textbf{34.41} & \textbf{21.52}  \\
        \bottomrule
    \end{tabular}
    \label{tab:tab2}
\end{table}

\subsection{Ablation Studies}
We conduct ablation studies on the nuScenes validation set to evaluate the effectiveness of the proposed HGFA module. All experiments adopt VGGT \cite{wang2025vggt} as the pre-trained VFM, forming the baseline configuration denoted as VG3S-base.

\subsubsection{Effect of the HGFA}
To assess the necessity of the proposed feature adapter, we replace the entire HGFA module with a standard DPT layer \cite{DPT}. As reported in Table \ref{tab:tab3}, this substitution causes a severe performance drop, yielding only 30.59\% in IoU and 19.31\% in mIoU. This substantial degradation demonstrates that a naive integration of frozen VFM tokens is insufficient for the downstream task. In contrast, the HGFA effectively adapts the generic VFM embeddings and unlocks their 3D geometric grounding capability, leading to significantly improved semantic occupancy prediction.

\begin{table}[tbp]
    \centering
    \caption{Effect of the HGFA.}
    \label{tab:tab3}
    \begin{tabular}{l|c|cc}
        \toprule
       Method & HGFA & IoU & mIoU \\
        \midrule    
        VG3S-base & \ding{51} & \textbf{33.29}  & \textbf{21.10} \\
        w/o HGFA & \ding{55} & 30.59 & 19.31 \\
        \bottomrule
    \end{tabular}
\end{table}

\begin{table}[tbp]
    \centering
    \caption{Ablation study on the components of the HGFA.}
    \label{tab:tab4}
    \begin{tabular}{l|ccc|cc}
        \toprule
        Method & GATF & TATR & LSFP & IoU & mIoU \\
        \midrule
        VG3S-base & \ding{51} & \ding{51} & \ding{51} & \textbf{33.29}  & \textbf{21.10} \\
        w/o GATF & \ding{55} & \ding{51} & \ding{51} & 32.52 & 20.43 \\
        w/o TATR & \ding{51} & \ding{55} & \ding{51} & 32.70 & 20.38 \\
        w/o LSFP & \ding{51} & \ding{51} & \ding{55} & 32.57 & 20.33 \\
        \bottomrule
    \end{tabular}
    \vspace{-0.25cm}
\end{table}

\begin{table}[t]
    \centering
    \caption{Effect of the grouping strategy in the GATF.}
    \label{tab:tab5}
    \begin{tabular}{c|cc}
        \toprule
        \# of groups ($K$) &  IoU &  mIoU  \\
        \midrule
        1  & 31.45 & 19.52 \\
        3 & 32.66 & 20.25 \\
        4  & \textbf{33.29} & \textbf{21.10} \\
        6  & 32.57 & 20.55 \\
        \bottomrule
    \end{tabular}
    \vspace{-0.35cm}
\end{table}

\subsubsection{Effect of Components in the HGFA}
Table \ref{tab:tab4} presents ablation results demonstrating the individual contributions of the constituent modules within the HGFA pipeline. We systematically disable or modify each module while keeping the rest of the architecture unchanged.
First, removing the GATF module causes a performance drop of 2.3\% in IoU. This decline indicates that the GATF is indispensable for effectively fusing intra-group features and generating the highly informative representations required for dense occupancy prediction.
Second, disabling the TATR module yields decreases of 1.8\% in IoU and 3.4\% in mIoU. This degradation highlights the importance of the TATR in calibrating and aligning VFM features with the requirements of the downstream task.
Finally, to isolate the impact of the LSFP, we replace it with a naive linear interpolation operation. Without the dedicated LSFP design, performance drops to 32.57\% in IoU and 20.33\% in mIoU, highlighting that the LSFP effectively enhances spatial correspondence and preserves local geometric consistency through its multi-scale reconstruction mechanism. Collectively, these results confirm that each component of the HGFA contributes positively to the exploitation of the cross-view 3D geometric grounding capability embedded in VFMs, ultimately improving overall semantic occupancy prediction performance.

\subsubsection{Effect of the Grouping Strategy in the GATF}
Further, we study the impact of the token grouping strategy within the GATF by varying the number of groups $K$, including $K \in \{1, 3,4,6\}$, where $K=1$ corresponds to no grouping. As shown in Table \ref{tab:tab5}, the configuration with $K=4$ achieves the best performance. This indicates that partitioning into four groups ensures tokens within each group share similar semantic granularity, allowing for effective aggregation and abstraction. This specific configuration strikes an optimal balance between structural fidelity and information compression, maximizing the utility of the frozen VFM tokens and subsequently boosting occupancy prediction accuracy.

\subsection{Qualitative Results}
We present qualitative comparisons between our proposed VG3S and the baseline GaussianFormer-2 \cite{huang2025gaussianformer2} across four representative driving scenes, as presented in Fig.~\ref{fig:fig3}. For clearer comparison, planar regions (e.g., drivable surfaces) are highlighted with boxes, while structural objects (e.g., buildings in the manmade category) are marked with circles. In the first two scenes, VG3S generates significantly more continuous and coherent road surfaces compared to GaussianFormer-2, resulting in superior reconstruction of the ground plane. In the third scene, the baseline struggles to accurately predict the parking lot on the left and the building structures on the right. Similarly, in the final intersection scenario, it fails to capture the upper building and yields a highly incomplete structure around the lower intersection. In contrast, VG3S consistently preserves structural integrity, faithfully recovering complete object geometries and dense scene layouts across these challenging environments. These visualizations demonstrate that VG3S effectively enables the solid cross-view geometric grounding from VFMs to substantially elevate occupancy prediction performance. More visualization results are provided in the supplementary video.

\section{Conclusion}
In this paper, we present VG3S, a generic and flexible framework that advances 3D semantic occupancy prediction by leveraging the powerful cross-view geometric grounding capability of VFMs. To this end, we design a plug-and-play hierarchical geometric feature adapter that injects rich 3D geometric priors from frozen VFMs into the Gaussian-based occupancy prediction pipeline. Specifically, VG3S employs the GATF to aggregate hierarchical VFM embeddings, the TATR to perform task-oriented feature alignment, and the LSFP to enable multi-scale spatial restructuring. Extensive experiments on the nuScenes occupancy prediction benchmark demonstrate that VG3S effectively exploits the solid 3D geometric priors embedded within diverse VFMs, yielding substantial improvements over the baseline and highlighting the immense promise of VFM-derived geometric grounding for advancing 3D semantic scene understanding.


\bibliography{reference}
\end{document}